\documentclass[10.5pt,compsoc]{TsT}
\usepackage{graphicx}
\usepackage{footmisc}
\usepackage{subfigure}
\usepackage{url}
\usepackage{multirow}
\usepackage[noadjust]{cite}
\usepackage{amsmath,amsthm}
\usepackage{amssymb,amsfonts}
\usepackage{booktabs}
\usepackage{color}
\usepackage{ccaption}
\usepackage{booktabs}
\usepackage{float}
\usepackage{fancyhdr}
\usepackage{caption}
\usepackage{xcolor,stfloats}
\usepackage{comment}
\graphicspath{{figures/}}
\usepackage{cuted}  
\usepackage{captionhack}
\usepackage{epstopdf}

\usepackage{color, colortbl, xcolor}
\usepackage{algorithmic}
\usepackage{algorithm}

\usepackage{tabularx}
\usepackage{todonotes}
\usepackage{cleveref}
\usepackage{url}
\usepackage{enumitem}

\newcommand{\hide}[1]{}
\newcommand{\mat}[1]{{\bf #1}}

\headevenname{\zihao{-5}{\textbf{\emph{Tsinghua Science and Technology, June}}} 2013, 18(3): 000-000}%
\headoddname{{\sf First author et al.:}\quad {\textbf{\emph{Click and Type Your Title Here ...}}}}%

\newtheoremstyle{mystyle}{0pt}{0pt}{\normalfont}{1em}{\bf}{}{1em}{}
\theoremstyle{mystyle}
\renewcommand\figurename{Fig.~}

\makeatletter
\renewcommand{\@biblabel}[1]{[#1]\hfill}
\makeatother

\begin{document}

\thispagestyle{empty}

\begin{strip}\zihao{3}
\noindent
\\ \textbf{Template for Preparation of Manuscripts for \\ \emph{Tsinghua Science and Technology}}
\vskip 6mm
\zihao{5}

\noindent
This template is to be used for preparing manuscripts for submission to \emph{Tsinghua Science and Technology}. Use of this template will save time in the review and production processes and will expedite publication. However, use of the template is not a requirement of submission. Do not modify the template in any way (delete spaces, modify font size/line height, etc.).
\vspace{180mm}
\end{strip}
\clearpage

\hyphenpenalty=50000

\makeatletter
\newcommand\mysmall{\@setfontsize\mysmall{7}{9.5}}

\newenvironment{tablehere}
  {\def\@captype{table}}
  {}
\newenvironment{figurehere}
  {\def\@captype{figure}}
  {}

\thispagestyle{plain}%
\thispagestyle{empty}%

\let\temp\footnote
\renewcommand \footnote[1]{\temp{\zihao{-5}#1}}
{}
\vspace*{-40pt}

\noindent{\zihao{5-}\textbf{\scalebox{0.89}[1.0]{\makebox[5.6cm][s]{%
TSINGHUA SCIENCE AND TECHNOLOGY}}}}

\vskip .2mm
{\zihao{5-}
\textbf{
\hspace{-5mm}
\scalebox{1}[1.0]{\makebox[5.6cm][s]{%
I\hspace{0.70pt}S\hspace{0.70pt}S\hspace{0.70pt}N\hspace{0.70pt}{\color{white}%
l\hspace{0.70pt}l\hspace{0.70pt}}1\hspace{0.70pt}0\hspace{0.70pt}0\hspace{0.70pt%
}7\hspace{0.70pt}-\hspace{0.70pt}0\hspace{0.70pt}2\hspace{0.70pt}1\hspace{0.70pt%
}4\hspace{0.70pt}{\color{white}l\hspace{0.70pt}l\hspace{0.70pt}}0\hspace{0.70pt}%
?\hspace{0.70pt}/\hspace{0.70pt}?\hspace{0.70pt}?\hspace{0.70pt}{\color{white}%
l\hspace{0.70pt}l\hspace{0.70pt}}p\hspace{0.70pt}p\hspace{0.70pt}?\hspace{0.70pt}?\hspace{0.70pt}?%
-\hspace{ 0.70pt}?\hspace{0.70pt}?\hspace{0.70pt}?}}}

\vskip .2mm\noindent
{\zihao{5-}\textbf{\scalebox{1}[1.0]{\makebox[5.6cm][s]{%
V\hspace{0.8pt}o\hspace{0.8pt}l\hspace{0.8pt}u\hspace{0.8pt}m\hspace{0.8pt}%
e\hspace{0.6em}1\hspace{0.8pt}8,\hspace{0.6em}N\hspace{0.8pt}u\hspace{0.8pt}%
m\hspace{0.8pt}b\hspace{0.8pt}e\hspace{0.8pt}r\hspace{0.6em}3,\hspace{0.6em}%
J\hspace{0.8pt}u\hspace{0.8pt}n\hspace{0.8pt}e%
\hspace{0.6em}2\hspace{0.8pt}0\hspace{0.8pt}1\hspace{0.8pt}3}}}}

\begin{strip}
{\center \vskip 3mm
{\zihao{3}\textbf{
Ranking with Adaptive Neighbors}}
\vskip 9mm}

{\center {\sf \zihao{5}
Muge Li, Liangyue Li, and Feiping Nie$^*$
}
\vskip 5mm}
%

\centering{
\begin{tabular}{p{160mm}}

{\zihao{-5}
\linespread{1.6667} %
\noindent
\bf{Abstract:} {\sf
Retrieving the most similar objects in a large-scale database for a given query is a fundamental building block in many application domains, ranging from web searches, visual, cross media, and document retrievals. State-of-the-art approaches have mainly focused on capturing the underlying geometry of the data manifolds. Graph-based approaches, in particular, define various diffusion processes on  weighted data graphs. Despite success, these approaches rely on  fixed-weight graphs, making ranking sensitive to the input affinity matrix. In this study, we propose a new ranking algorithm that simultaneously learns the data affinity matrix and the ranking scores. The proposed optimization formulation assigns adaptive neighbors to each  point in the data based on the local connectivity, and the smoothness constraint assigns similar ranking scores to similar data points. We develop a novel and efficient algorithm to solve the optimization problem. Evaluations using synthetic and real datasets suggest that the proposed algorithm can outperform the existing methods. }
}

\end{tabular}
}
\vskip 6mm

\vskip -3mm
\zihao{6}\end{strip}

\thispagestyle{plain}%
\thispagestyle{empty}%
\makeatother
\pagestyle{tstheadings}

\begin{figure}[b]
\vskip -6mm
\begin{tabular}{p{44mm}}
\toprule\\
\end{tabular}
\vskip -4.5mm
\noindent
\setlength{\tabcolsep}{1pt}
\begin{tabular}{p{1.5mm}p{79.5mm}}
$\bullet$& Muge Li is with Cixi Hanvos Yucai High School, Ningbo, China, 315300. E-mail: 1606024250@qq.com.\\
$\bullet$& Liangyue Li is with with the School of Computing, Informatics, Decision Systems Engineering, Arizona State University, Tempe, AZ, US, 85281. E-mail: liangyue@asu.edu.\\
$\bullet$ & Feiping Nie is with the School of Computer Science and Center for OPTical IMagery Analysis and Learning (OPTIMAL), Northwestern Polytechnical University, Xi’an, China, 710072. E-mail: feipingnie@gmail.com.
 \\
$\sf{*}$&
To whom correspondence should be addressed. \\
          &          Manuscript received: year-month-day; revised: year-month-day; accepted: year-month-day

\end{tabular}
\end{figure}\zihao{5}

\vbox{}
\vskip 1mm
\noindent

\section{Introduction}\label{sec:intro}
\noindent
Retrieving the most similar objects in a large-scale database for a given query is a fundamental building block in many application domains, ranging from web search~\cite{page1999pagerank}, visual retrieval~\cite{he2004manifold,tong2006manifold,bai2017regularized,donoser2013diffusion,iscenefficient}, cross media retrieval~\cite{yang2009ranking}, to document retrieval~\cite{Cao:2006:ARS:1148170.1148205}. The most straightforward approach to such retrieval tasks is to compute the pairwise similarities between objects in the Euclidean space as the ranking scores. Nonetheless, high-dimensional data often lie on a nonlinear manifold~\cite{roweis2000nonlinear,tenenbaum2000global}. The Euclidean distance based approach largely ignores the intrinsic manifold structure and might degrade the retrieval performance. 

State-of-the-art methods mainly focus on capturing the underlying geometry of the data manifold. The most common way is to first represent the data manifold using a weighted graph, wherein each vertex is a data object, and the edge weights are proportional to the pairwise similarities. All the vertices then repeatedly spread their affinities to their neighborhood via the weighted graph until a global stable state is reached. The various diffusion processes mainly differ in the transition matrix and the affinity update scheme~\cite{donoser2013diffusion}. Among others, the random walk transition matrix is widely used in PageRank~\cite{page1999pagerank}, random walk with restart~\cite{tong2006fast}, self diffusion~\cite{wang2012affinity}, label propagation~\cite{zhu2003semi} and graph transduction~\cite{bai2010learning}. The random walk transition matrix is a row-stochastic matrix such that the transition probability is proportional to the edge weights.A slight variant is the symmetric normalized transition matrix used in the Ranking on Data Manifold method~\cite{Zhou:2003:RDM:2981345.2981367}. To reduce the effect of noisy nodes, random walks can be restricted to the $k$ nearest neighbors by sparsifying the original weighted graph~\cite{Szummer:2001:PLC:2980539.2980661,5206844}.
For iterative update of the affinities, the random walk with restart allows for the random surfer to randomly jump to an arbitrary node. The modified diffusion process on the standard graph captures the high-order relations~\cite{5206844} and is equivalent to the diffusion process on the Kronecker product graph~\cite{yang2013affinity}. Despite success,  graph-based ranking methods rely on  fixed-weight graphs, making the ranking results sensitive to the input affinity matrix. 

In this study, we propose the ranking with adaptive neighbors (RAN) algorithm simultaneously learns the data affinity matrix and the ranking scores. The proposed optimization explores  two objectives. First, data points with smaller distance in the Euclidean space have high chance to be neighbors, i.e., more similar. In contrast to other graph-based ranking methods, the similarity is not computed a priori but is learned via optimizing the ranking scores. Consequently, the neighbors of each datum are adaptively  assigned. Second, similar data points have similar ranking scores. This is essentially the smoothness constraint in  graph transduction methods~\cite{wang2008graph}. We develop a novel and efficient algorithm to solve the optimization problem. Evaluations using synthetic and real datasets suggest that the proposed ranking algorithm outperforms  existing methods.

In section~\ref{sec:alg}, we present the proposed RAN algorithm. Next, in section~\ref{sec:exp} we discuss the empirical evaluation results and, in section~\ref{sec:conclusions}, we summarize the conclusions.

{\bf Notations:} Throughout the paper, the matrices are written as upper-case letters. For matrix $M$, the $i$-th row and  $(i,j)$-th element of $M$ are denoted by $m_i$ and $m_{ij}$, respectively. An identity matrix is denoted by $I$, and $\mat{1}$ denotes the column vector with all elements as one. For vector $v$ and matrix $M$, $v\ge 0$ and $M \ge 0$ represent all the elements of $v$ and $M$ are nonnegative.

\section{Ranking with Adaptive Neighbors}\label{sec:alg}
\noindent
In this section, we discuss RAN algorithm and then the optimization approach for solving the objective function.

\subsection{Proposed Formulation}
\noindent
Given a set of data points $\mathcal{X} = \{x_1, x_2, \ldots, x_N\} \subseteq \mathbb{R}^d$ with a query indicator vector $y=[y_1, y_2, \ldots, y_N]^T \in \{0,1\}^N$, where $y_1=1$ if $x_i$ is the query and $y_1=0$ otherwise, the task is to find a function $f$ that assigns each point in the data $x_i$ a ranking score $f_i \in \mathbb{R}$ according to its relevance to the queries. We explore the local connectivity of each  point for ranking purposes and in particular consider the $k$-nearest  points as the neighbors of a specific node. 

Data points separated by small distances in the Euclidean space have high chance to be neighbors. We denote the probability that the $i$-th data point $x_i$, and the $j$-th data point $x_j$ are neighbors by $s_{ij}$. Intuitively, if the two data points are separated by a small distance, i.e.,  $\|x_i - x_j\|^2_2$ is small, then their probability $s_{ij}$ of being connected is likely high. One way to find such probabilities $s_{ij}|^N_{j=1}$ is to solve the following optimization problem:
\begin{equation}
\min_{s_i^T\mat{1} = 1, 0\le s_i\le 1} \sum_{j=1}^N \|x_i - x_j\|^2_2 s_{ij}
\end{equation}
\noindent where $s_i\in \mathbb{R}^{N}$ is a vector with the $j$-th element as $s_{ij}$. Nonetheless, the above optimization problem has a trivial solution, that is, $s_{ij} =1$ for the nearest data point $x_j$ of $x_i$, otherwise  $s_{ij} =0$. This can be addressed by adding a $l_2$-norm  regularization on $s_{i}$ to drag $s_i$ closer to the center of mass of the simplex defined by $s_i^T\mat{1} = 1, 0\le s_i\le 1$. This slight modification gives us the following optimization problem:
\begin{equation}\label{eq:sij}
\min_{s_i^T\mat{1} = 1, 0\le s_i\le 1} \sum_{j=1}^N (\|x_i - x_j\|^2_2 s_{ij} + \gamma s_{ij}^2)
\end{equation}
\noindent where the second term is the regularization term and $\gamma$ is the regularization parameter. 

For each data point $x_i$, we  compute its probability of connecting to other data points using Eq.~\eqref{eq:sij}. As a result, we  assign the neighbors of all the data points by solving the following problem:
\begin{equation}
\min_{\forall i, s_i^T\mat{1} = 1, 0\le s_i\le 1} \sum_{i,j=1}^N (\|x_i - x_j\|^2_2 s_{ij} + \gamma s_{ij}^2)
\end{equation}

Similar data points have similar ranking scores, essentially a smoothness constraint over the data graph. We assume the matrix $S\in \mathbb{R}^{N\times N}$ is the similarity matrix obtained from assigning the neighbors, where each row is $s_i^T$. We  write the smoothness constraint as,
\begin{equation}\label{eq:fLf}
\sum_{i,j=1}^N (f_i -f_j)^2 s_{ij} = 2 f^TL_Sf
\end{equation}
\noindent where $f$ is the vector of ranking scores for all the data points, $L_S = D_S - \frac{S^T + S}{2}$ is the Laplacian matrix of the affinity matrix, and the degree matrix $D_S$ is a diagonal matrix with the $i$-th diagonal element defined as $\sum_j (s_{ij} + s_{ji})/2$.

Combining the above and using the information from the query, we derive the final objective function:
\begin{equation}\label{eq:objective}
\begin{split}
 \min_{S,f} &\sum_{i,j=1}^n (\|x_i - x_j\|_2^2 s_{ij} + \gamma s_{ij}^2) + 2\lambda f^TL_S f \\
 &+ (f-y)^T U (f-y) \\
 s.t. & \quad  \forall i, s_i^T \mat{1} = 1, 0\leq s_i\leq 1
\end{split}
\end{equation}

\noindent where $U$ is a diagonal matrix with $U_{ii} = \infty$ (a large constant) if $x_i$ is the query, otherwise $U_{ii} = 1$. The last term is equivalent to $\sum_{i=1}^n U_{ii}(f_i - y_i)^2$ to make  the ranking results consistent with the queries. The queries are given much more weights as they reflect the user's search intentions. In non-queried examples, we do not know a priori whether they meet the user's intentions and give them lower weights. It is not easy to solve Eq.~\eqref{eq:objective} because $L_S = D_S - \frac{S^T + S}{2}$ and $D_S$ both depend on the similarity matrix $S$. In the next subsection, we  propose a novel and efficient algorithm to solve this  problem. 

\subsection{Optimization Solutions}
\noindent
We propose to solve Eq.~\eqref{eq:objective} via an alternative optimization approach. We first fix $S$ and then the problem transforms to:
\begin{equation}
\min_f 2\lambda f^T L_S f + (f- y)^T U (f - y)
\end{equation}

We take the derivative of the above objective function w.r.t. $f$ and set it to 0, obtaining the following linear equation:
\begin{equation}\label{eq:solve_f}
 (2\lambda L_S + U)f = Uy
 \end{equation}
 
 The solution is easily obtained as $f = (2\lambda L_S + U)^{-1}Uy$.
 
 When $f$ is fixed, Eq.~\eqref{eq:objective} transforms to:
  \begin{align}
 & \min_{S} \sum_{i,j=1}^n (\|x_i - x_j\|_2^2 s_{ij} + \gamma s_{ij}^2) + 2\lambda f^TL_S f \\
 & s.t.  \quad  \forall i, s_i^T \mat{1} = 1, 0\leq s_i\leq 1
 \end{align}
 
And based on Eq.~\eqref{eq:fLf}, it is written
\begin{equation}
   \begin{split}
  \min_{S} &\sum_{i,j=1}^n (\|x_i - x_j\|_2^2 s_{ij} + \gamma s_{ij}^2 + \lambda (f_i - f_j)^2s_{ij} )\\
  s.t.  &  \forall i, s_i^T \mat{1} = 1, 0\leq s_i\leq 1
 \end{split}
 \end{equation}
 Because the summations are independent of each other given $i$, we can solve the following sub-problem individually for each $i$:
 \begin{equation}\label{eq:subprob}
  \begin{split}
 \min_{s_i} & \sum_{j=1}^n (\|x_i - x_j\|_2^2 s_{ij} + \gamma s_{ij}^2 + \lambda (f_i - f_j)^2s_{ij} )\\
 s.t. &    s_i^T \mat{1} = 1, 0\leq s_i\leq 1
 \end{split}
  \end{equation}

 We denote $d_{ij}^x = \|x_i - x_j\|_2^2$ and $d_{ij}^f = (f_i - f_j)^2$, and denote $d_i \in \mathbb{R}^{N}$ as a vector with the $j$-th element as $d_{ij} = d_{ij}^x + \lambda d_{ij}^f$. Then Eq.~\eqref{eq:subprob} is reformulated as:
 \begin{equation}\label{eq:solve_si}
 \min_{s_i^T \mat{1} = 1, 0\leq s_i\leq 1}\|s_i + \frac{d_i}{2\gamma} \|_2^2
 \end{equation}
 
Next, we will show how to solve this equation in a closed form using the Lagrange multipliers method. The Lagrangian function of the problem is 
 \begin{equation}
 \mathcal{L}(s_i, \eta, \beta_i) = \frac{1}{2}\|s_i + \frac{d_i}{2\gamma_i} \|_2^2 - \eta(s_i^T\mat{1} - 1) - \beta_i^Ts_i
 \end{equation}
 \noindent where $\eta$ and $\beta_i$ are  non-negative Lagrangian multipliers. 
 
 According to the KKT condition, the optimal solution is 
 \begin{equation}\label{eq:s_ij}
 s_{ij} = (- \frac{d_{ij}}{2\gamma_i} + \eta)_{+}
 \end{equation}
 \noindent where  $(x)_+$ is the shorthand for $\max\{x,0\}$.
 
 It is often desirable to focus on the locality of each  point, as it can reduce the effect of noisy data  and boost the performance in practice~\cite{Nie:2014:CPC:2623330.2623726}. In this study, we will learn the sparse vector $s_i$ and allow $x_i$ to connect to its $k$-nearest neighbors. Such sparsification of $S$ would minimize the computational cost. 
 
We sort $d_{ij}$ in ascending order such that $d_{i1} \le d_{i2} \le \ldots \le d_{iN}$. We want to learn the sparse $s_i$ with only $k$ nonzero elements, from Eq.~\eqref{eq:s_ij}; thus we have $s_{ik}>0$ and $s_{i, k+1} = 0$. Therefore
 \begin{equation}\label{eq:sparse_s}
 \begin{cases}
    - \frac{d_{ik}}{2\gamma_i} + \eta > 0 \\
    - \frac{d_{i,k+1}}{2\gamma_i} + \eta \leq 0
\end{cases}
 \end{equation}
 
 Considering the constraint $s_i^T\mat{1} = 1$, we obtain
 \begin{equation}\label{eq:gamma}
 \sum_{j=1}^k (- \frac{d_{ij}}{2\gamma_i} + \eta) = 1 \Rightarrow \eta = \frac{1}{k} + \frac{1}{2k\gamma_i}\sum_{j=1}^k d_{ij}
 \end{equation}
 
 Substituting Eq.~\eqref{eq:gamma} into Eq.~\eqref{eq:sparse_s}, we obtain the following inequality for $\gamma_i$
 \begin{equation}
 \frac{k}{2} d_{ik} -\frac{1}{2}\sum_{j=1}^k d_{ij} < \gamma_i \leq \frac{k}{2}d_{i,k+1} - \frac{1}{2}\sum_{j=1}^k d_{ij}
 \end{equation}
 
 For the objective function in Eq.~\eqref{eq:solve_si} to have an optimal solution $s_i$, we  set $\gamma_i$ to 
 \begin{equation}
 \gamma_i = \frac{k}{2}d_{i,k+1} - \frac{1}{2}\sum_{j=1}^k d_{ij}
 \end{equation}
 
 The overall $\gamma$ is set as the mean of all $\gamma_i$: 
  \begin{equation}
 \gamma = \frac{1}{n} \sum_{i=1}^n (\frac{k}{2}d_{i,k+1} - \frac{1}{2}\sum_{j=1}^k d_{ij})
 \end{equation}
 
 The  algorithm for solving the optimization problem in Eq.~\eqref{eq:objective} is summarized in Algorithm~\ref{alg:overall}.
 
 \begin{algorithm}[!htb]
	\caption{Algorithm to solve problem in Eq.~\eqref{eq:objective}}\label{alg:overall}
	\begin{algorithmic}[1]
		\REQUIRE {(1) Data matrix $X\in \mathbb{R}^{n\times d}$, \\
        		  (2) Query indicator vector $y$,\\
				  (3) parameters $\gamma$, $\lambda$.}\\
				 
		\ENSURE {The ranking scores $f$. }
		\STATE Initialize $S$ and compute $L_S$ accordingly;
		\WHILE {not converged}
        \STATE Define the diagonal matrix $U$ as: $U_{ii} = \infty$ if $y_i=1$ and $U_{ii} = 1$ otherwise;
		\STATE Update $f$ by solving Eq.~\eqref{eq:solve_f} as $f = (2\lambda L_S + U)^{-1}Uy$;
        \FOR{$i=1,\ldots, N$}
         \STATE Update $i$-th row of $S$ by solving Eq.~\eqref{eq:solve_si}
        \ENDFOR
		\ENDWHILE
		
	\end{algorithmic}
\end{algorithm}

\hide{
\subsection{Existing Work}
Existing work~\cite{Zhou:2003:RDM:2981345.2981367}

 \begin{equation}
 \min_{f} f^T L f + (f-y)^T U (f-y) 
 \end{equation}
 \noindent where $L$ is the normalized Laplacian matrix and is defined as $L = I - D^{-1/2}WD^{-1/2}$ and $D$ is a diagonal matrix with the elements $D_{ii} = \sum_j W_{ij}$. The matrix $U$ is a diagonal matrix to assign different weights to different data points with $U_{ii} = \infty$ (a large constant) if $x_i$ is the query and $U_{ii} = 1$ otherwise.
 
 The limitation is that the data graph is fixed and the ranking results are thus sensitive to the input affinity matrix. 
 }

\section{Experiments}\label{sec:exp}
In this section, we show the performance of the proposed ranking algorithm RAN (Algorithm~\ref{alg:overall}) on synthetic and real world datasets.

\subsection{Synthetic datasets}
\noindent
We randomly generate two synthetic datasets constructed as two moons (Fig.~\ref{fig:twomoon}) and three rings (Fig.~\ref{fig:threering}) patterns. A query is given in the upper moon and the innermost ring marked in red cross. The task is to rank the remaining data points according to their relevance to the query. We represent the ranking scores returned by RAN using the diameter of the data points such that larger points are more relevant. From Fig.~\ref{fig:twomoon}, we observe that the ranking scores gradually decrease along the upper moon. The same decreasing trend is also observed in the lower moon. In addition, the ranking scores in the upper moon are generally much higher than in the lower moon. Such ranking outcome is intuitively expected. We make similar observations for the three rings  in Fig.~\ref{fig:threering}. The data points in the innermost ring are more relevant than those in the middle ring, which are more relevant than those in the outermost ring. These results clearly show that the proposed RAN can capture the underlying manifold pretty well. 

\begin{figure}[!htb]
\centering
\includegraphics[width=0.45\textwidth]{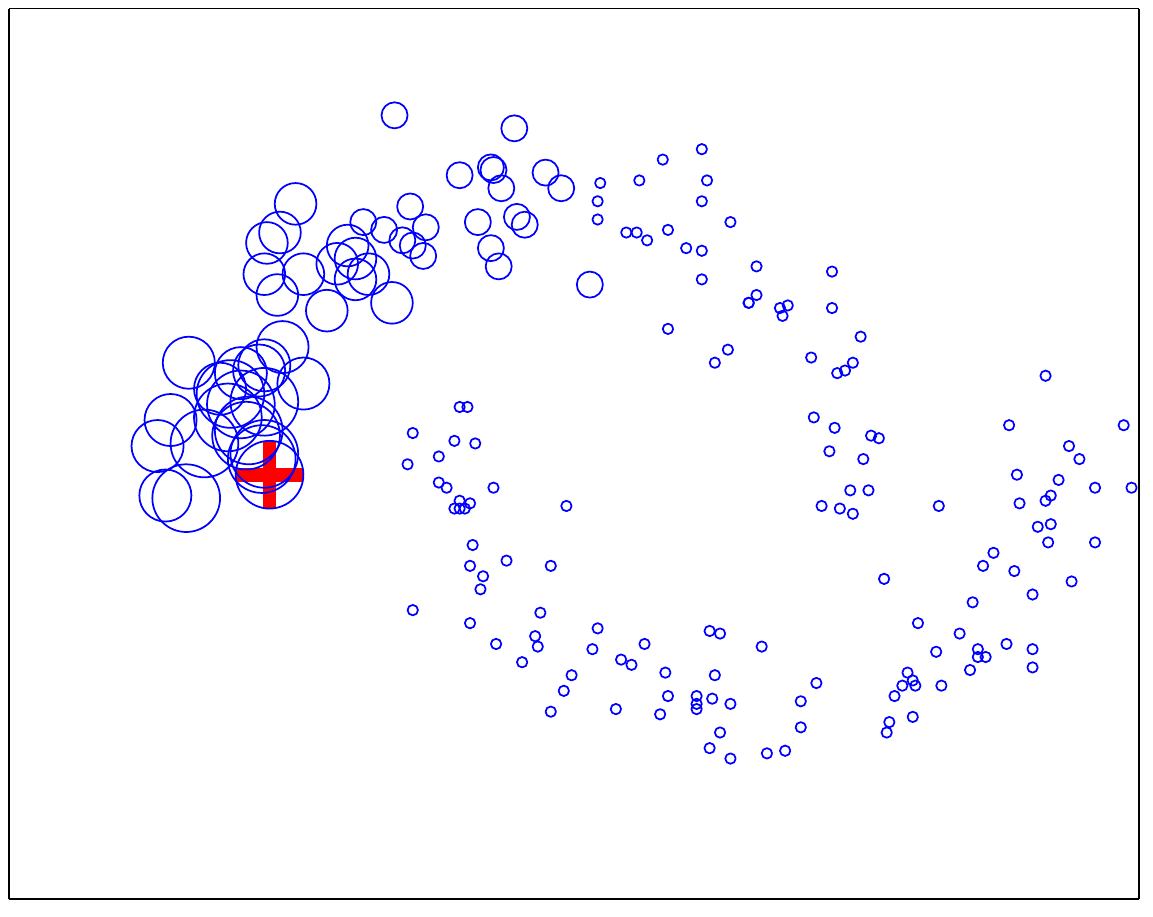}
\caption{Ranking Example using Two Moon.}
\label{fig:twomoon}
\end{figure}

\begin{figure}[!htb]
\centering
\includegraphics[width=0.45\textwidth]{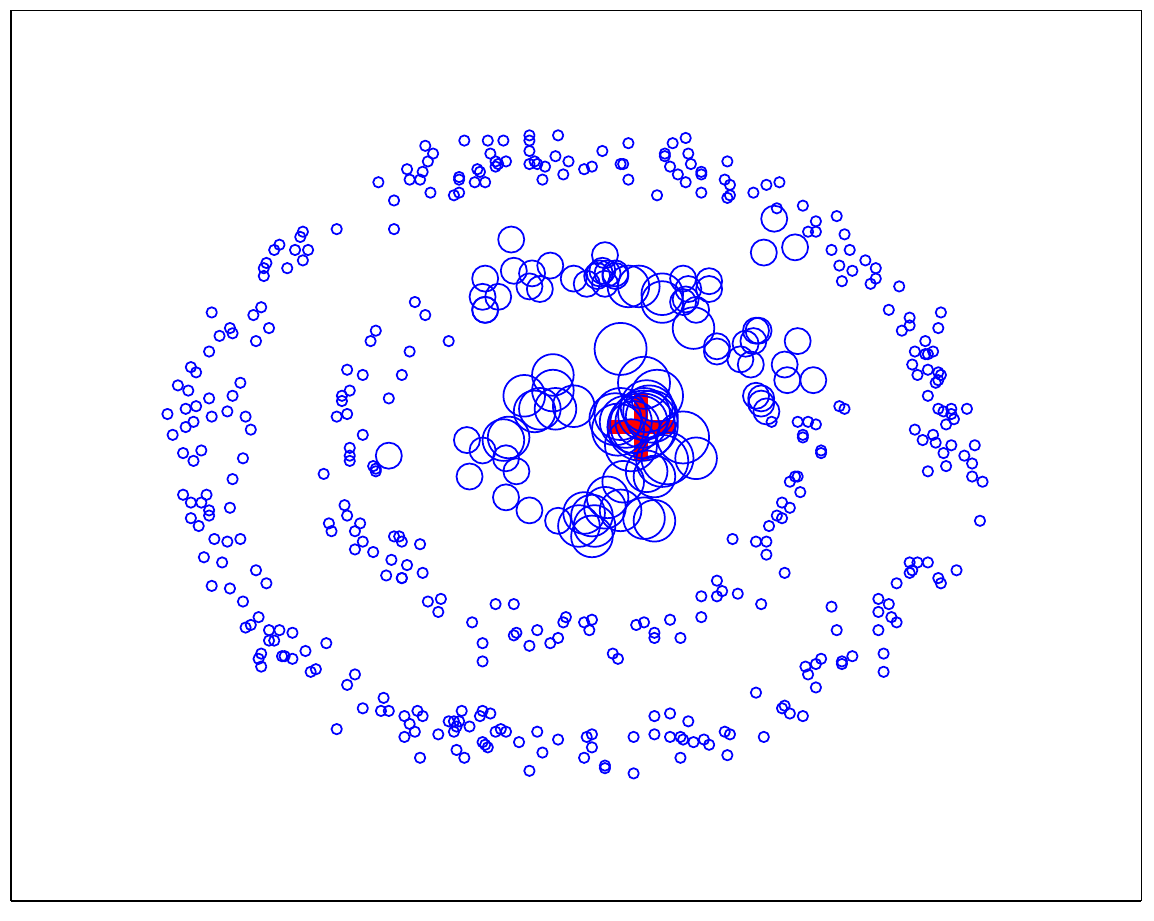}
\caption{Ranking Example using Three Ring.}
\label{fig:threering}
\end{figure}

\subsection{Real dataset}
\noindent
We  compare the retrieval performance on three real  image datasets: Yale~\cite{georghiades2001few}, ORL~\cite{samaria1994parameterisation} and USPS~\cite{hull1994database}.

{\bf YALE:} Yale contains face images of subjects at different poses and illumination conditions. We extract 11 images at different conditions for 15 subjects. Each image is down-sampled and normalized to zero mean and unit variance. The bandwidth for constructing the weighted graph for the graph based baselines is $\sigma = 0.021$. We set $k=5$ and $\lambda=90$ for RAN. 

{\bf ORL:} ORL contains  contains 400 images  with ten different images for 40 different subjects each. The bandwidth for constructing the weighted graph for the graph based baselines is $\sigma = 20$. We set $k=5$ and $\lambda=0.1$ for RAN.

{\bf USPS:}  This dataset collects images of handwritten digits (0-9)
from envelopes of the U.S. Postal Service. We extract 40 images for each digit and normalize them to 16 $\times$ 16 pixels in gray scale. The bandwidth for constructing the weighted graph for the graph based baselines is $\sigma = 0.8$. We set $k=10$ and $\lambda=1.0$ for RAN.

On all the datasets, we use each image as query and measure the retrieval accuracy by ranking all the other images. We compare the proposed RAN algorithm with the Euclidean distance based baseline and several other diffusion methods, including self-diffusion (SD)~\cite{wang2012affinity}, Personalized PageRank (PPR)~\cite{haveliwala2002topic}, Manifold Ranking~\cite{Zhou:2003:RDM:2981345.2981367} and Graph Transduction (GT)~\cite{bai2010learning}.  The results are shown in Tables~\ref{tab:yale},~\ref{tab:orl} and \ref{tab:usps}. From the results, we can see that the proposed RAN algorithm consistently outperforms all other methods. The straightforward Euclidean distance based baseline is the worst because it ignores the manifold structure in the data. The various diffusion based methods capture the manifold information to a certain extent, but they assume the weighted data graph is fixed. We instead adaptively learn the localized weighted graph optimized for the ranking. To study how the locality of the graph, i.e., the number of neighbors $k$, affects the retrieval performance, we show (Fig.~\ref{fig:neighbors}) the retrieval performance by varying the number of neighbors on USPS dataset. As it can be seen, it is important to select a reasonable value for $k$ for the retrieval. For USPS, the best performance can be achieved at $k=15$. 

\hide{
\begin{table}[!htb]
\caption{The comparison of retrieval performance (\%) on YALE. $\sigma = 0.1$ , $k=5$. $\sigma$ is the Gaussian kernel bandwidth for constructing the weight matrix.\\}
\centering
\begin{tabular}{|c|c|c|}
\hline
Methods & Precision@10 & Recall@10 \\
\hline
Euclidean Distance & 66.61 & 60.55 \\ 
\hline
SD~\cite{wang2012affinity} & 70.48 & 64.08 \\
\hline
PPR & 70.48 & 64.08\\
\hline
Manifold Ranking~\cite{Zhou:2003:RDM:2981345.2981367} & 70.48 & 64.08 \\
\hline
LCDP~\cite{yang2009locally} & 70.97& 64.52 \\
\hline
RAN (ours) & 71.94 & 65.40 \\
\hline
\end{tabular}
\end{table}
}

\begin{table}[!htb]
\caption{Retrieval performance (\%) for YALE.\\}
\vspace{2mm}
\centering
\begin{tabular}{|c|c|c|}
\hline
Methods & Precision@10 & Recall@10 \\
\hline
Euclidean Distance & 66.61 & 60.55 \\ 
\hline
SD~\cite{wang2012affinity} & 69.03 & 62.75 \\
\hline
PPR~\cite{haveliwala2002topic} & 69.03 & 62.75\\
\hline
Manifold Ranking~\cite{Zhou:2003:RDM:2981345.2981367} & 68.85 & 62.59 \\
\hline
GT~\cite{bai2010learning} & 68.91& 62.65 \\
\hline
RAN (ours) & {\bf 72.00} & {\bf 65.45} \\
\hline
\end{tabular}
\label{tab:yale}
\end{table}

\begin{table}[!htb]
\caption{Retrieval performance (\%) for ORL. \\}
\vspace{2mm}
\centering
\begin{tabular}{|c|c|c|}
\hline
Methods & Precision@15 & Recall@15 \\
\hline
Euclidean Distance & 41.56 & 62.35 \\ 
\hline
SD~\cite{wang2012affinity} & 46.87  & 70.30  \\
\hline
PPR~\cite{haveliwala2002topic} & 47.15 &  70.73\\
\hline
Manifold Ranking~\cite{Zhou:2003:RDM:2981345.2981367} & 47.35 & 71.02 \\
\hline
GT~\cite{bai2010learning} &  48.97 & 73.45 \\
\hline
RAN (ours) &  {\bf 49.02} & {\bf 73.53} \\
\hline
\end{tabular}
\label{tab:orl}
\end{table}

\begin{table}[!htb]
\caption{Retrieval performance (\%) for USPS.\\}
\vspace{2mm}
\centering
\begin{tabular}{|c|c|c|}
\hline
Methods & Precision@50 & Recall@50 \\
\hline
Euclidean Distance & 45.53 & 56.91 \\ 
\hline
SD~\cite{wang2012affinity} & 47.42   & 59.27  \\
\hline
PPR~\cite{haveliwala2002topic} & 47.39 &  59.24\\
\hline
Manifold Ranking~\cite{Zhou:2003:RDM:2981345.2981367} & 47.42 & 59.28 \\
\hline
GT~\cite{bai2010learning} &  46.18 & 57.72 \\
\hline
RAN (ours) &  {\bf 56.19} & {\bf 70.23} \\
\hline
\end{tabular}
\label{tab:usps}
\end{table}

\begin{figure}
\centering
\includegraphics[width=0.5\textwidth]{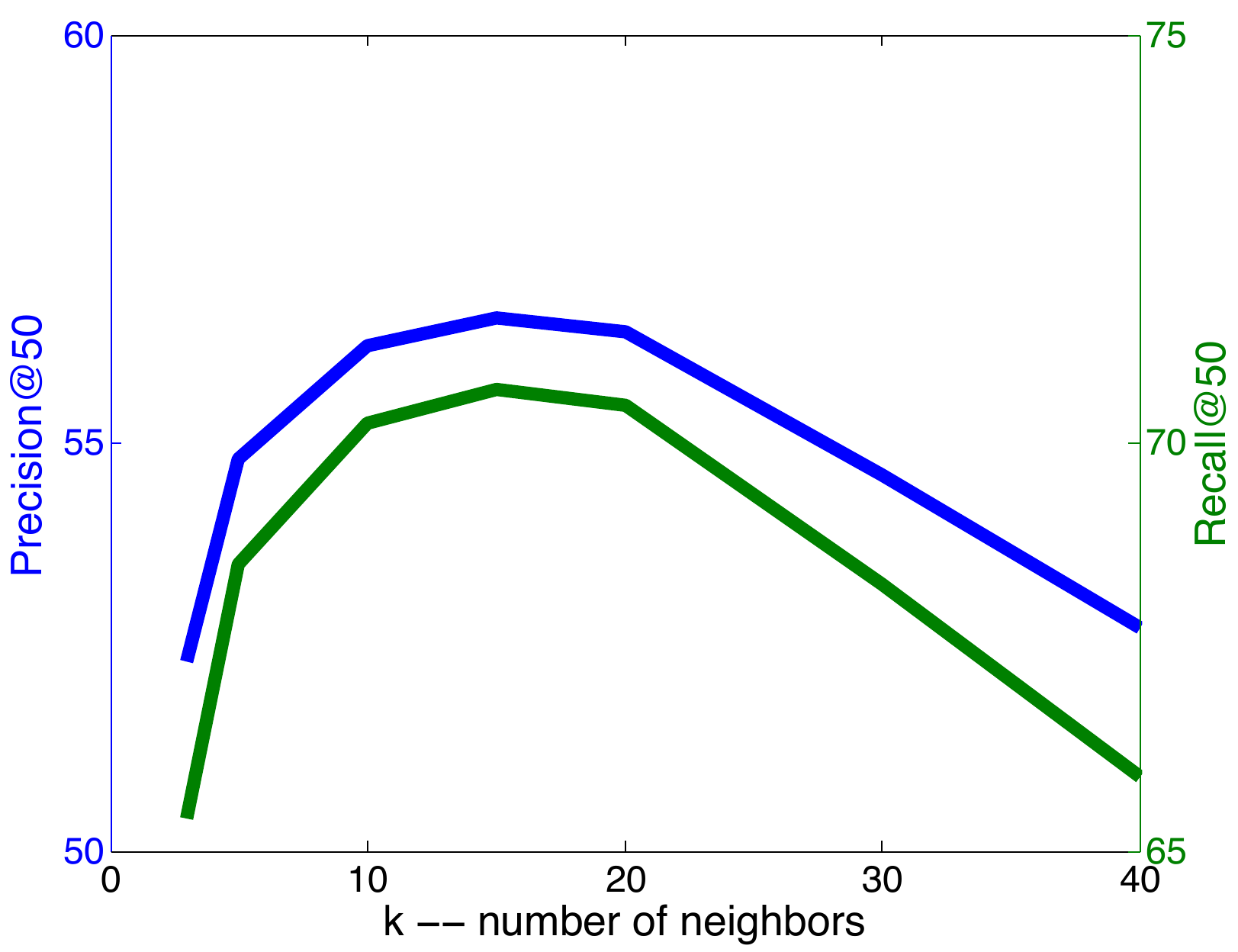}
\caption{Retrieval Performance (\%) v.s. the number of neighbors on USPS.}
\vspace{-4pt}
\label{fig:neighbors}
\end{figure}


\section{Conclusions}\label{sec:conclusions}
We study the data ranking problem by capturing the underlying geometry of the data manifold. Instead of relying on the fixed-weight data graphs, we propose a new ranking algorithm that is able to learn the data affinity matrix and the ranking scores simultaneously. The proposed optimization formulation assigns adaptive neighbors to each data point based on the local connectivity and the smoothness constraint assigns similar ranking scores to similar data points. An efficient algorithm is developed to solve the optimization problem. Evaluations using synthetic and real datasets demonstrates the superior performance of the proposed algorithm. 

\bibliographystyle{plain}
\bibliography{sigproc} 
\begin{strip}
\end{strip}

\begin{biography}[yourphotofilename.eps]
\noindent
\textbf{First B. Author}\ \  Photo. Biographies should be limited to one paragraph consisting of the following: sequentially ordered list of degrees, including years achieved; sequentially ordered places of employ concluding with current employment; associa-tion with any official journals or conferences; major profes-sional and/or academic achievements, i.e., best paper awards, research grants, etc.; any publication information (number of papers and titles of books published); current research interests; association with any professional associations.
\end{biography}

\begin{biography}[yourphotofilename.eps]
\noindent
\textbf{Second B. Author} Photo. Biographies should be limited to one paragraph consisting of the following: sequentially ordered list of degrees, including years achieved; sequentially ordered places of employ concluding with current employment; associa-tion with any official journals or conferences; major profes-sional and/or academic achievements, i.e., best paper awards, research grants, etc.; any publication information (number of papers and titles of books published); current research interests; association with any professional associations.
\end{biography}
\vskip 22mm
\begin{biography}[yourphotofilename.eps]
\noindent
\textbf{Third C. Author}  Photo. Biographies should be limited to one paragraph consisting of the following: sequentially ordered list of degrees, including years achieved; sequentially ordered places of employ concluding with current employment; associa-tion with any official journals or conferences; major profes-sional and/or academic achievements, i.e., best paper awards, research grants, etc.; any publication information (number of papers and titles of books published); current research interests; association with any professional associations.
\end{biography}

\begin{strip}
\end{strip}

\mbox{}
\clearpage
\clearpage
\end{document}